\documentclass[letterpaper, 10 pt, journal, twoside]{ieeeconf}   

\IEEEoverridecommandlockouts                              

\usepackage{amsmath}
\usepackage{yfonts}
\usepackage{booktabs}
\usepackage{amssymb}
\usepackage{graphicx}
\usepackage{multirow}
\usepackage{color}
\usepackage{float}
\usepackage{units}
\usepackage{footnote}
\usepackage{bm}
\makesavenoteenv{tabular}
\usepackage{pbox}
\makesavenoteenv{table}
\usepackage[all=normal,paragraphs=normal,floats=tight,mathspacing=normal,wordspacing=tight,charwidths=tight,mathdisplays=normal,leading=normal]{savetrees}
\newcommand{\camposetime}[1]{^{c_{#1}}\mathbf{T}_w}
\newcommand{\campose}[0]{^c\mathbf{T}_w}
\newcommand{\objposetime}[1]{^w\mathbf{T}_{o_{#1}}}
\newcommand{\objpose}[0]{^w\mathbf{T}_{o}}

\newcommand{\pointidx}[2]{^{#1}\mathbf{X}_{#2}}
\newcommand{\point}[1]{{^{#1}\mathbf{X}}}

\newcommand{\imgpointij}[2]{^{#1}\mathbf{x}_{#2}}
\newcommand{\twistcoord}[2]{^{#2}\bm{\xi}_{#1}}

\newcommand{\tildetwistcoord}[2]{^{#2}\bm{\Tilde{\xi}}_{#1}}
\newcommand{\twist}[0]{\bm{\xi}}
\newcommand{\pose}[2]{^{#1}\mathbf{T}_{#2}}
\newcommand{\coordframe}[1]{\mathcal{F}_{#1}}
\newcommand{\rot}[2]{^{#1}\mathbf{R}_{#2}}
\newcommand{\trans}[2]{^{#1}\mathbf{t}_{#2}}
\newcommand{\adj}[2]{^{#1}\mathbf{V}_{#2}}
\newcommand{\tspeed}[0]{\mathbf{v}}
\newcommand{\rspeed}[0]{\bm{\omega}}
\newcommand{\projmat}[0]{\bm{\Pi}}
\newcommand{\proj}[0]{p}

\newcommand{\rev}[0]{}

\title{\LARGE \bf
TwistSLAM: Constrained SLAM in Dynamic Environment 
}

\author{Mathieu Gonzalez$^{1}$, Eric Marchand$^{2}$, Amine Kacete$^{1}$ and Jerome Royan$^{1}$
\thanks{$^{1}$ Mathieu Gonzalez, Amine Kacete and Jerome Royan are with the Institute of Research and Technology b$<>$com, Rennes, France,
       {\tt\small \{mathieu.gonzalez,amine.kacete, jerome.royan\}@b-com.com}}%
\thanks{$^{2}$ Eric Marchand is with Univ Rennes, Inria, IRISA, CNRS, Rennes, France,
        {\tt\small Eric.Marchand@irisa.fr}}%
}

\usepackage{hyperref}
\begin{document}
\onecolumn
\begin{center}
This paper has been accepted for publication in \textit{IEEE Robotics and Automation Letters}.
~\\
~\\
~\\
DOI: \href{https://dx.doi.org/10.1109/LRA.2022.3178150}{10.1109/LRA.2022.3178150} 
~\\
IEEE Xplore: \href{https://ieeexplore.ieee.org/document/9782519}{https://ieeexplore.ieee.org/document/9782519}
~\\
~\\
~\\
~\\
© 2022 IEEE.  Personal use of this material is permitted.  Permission from IEEE must be obtained for all other uses, in any current or future media, including reprinting/republishing this material for advertising or promotional purposes, creating new collective works, for resale or redistribution to servers or lists, or reuse of any copyrighted component of this work in other works.
\end{center}
\twocolumn
\maketitle

\begin{abstract}
\rev{
Classical visual simultaneous localization and mapping (SLAM)  algorithms usually assume the environment to be rigid. This assumption limits the applicability of those algorithms as they are unable to accurately estimate the camera poses and world structure in real life scenes containing moving objects  (e.g. cars, bikes, pedestrians, etc.). To tackle this issue, we propose TwistSLAM:  a semantic, dynamic and stereo SLAM system that can track dynamic objects in the environment. Our algorithm creates clusters of points according to their semantic class.
Thanks to the definition of inter-cluster constraints modeled by mechanical joints (function of the semantic class), a novel constrained bundle adjustment  is then able to jointly estimate both poses and velocities of moving objects along with the classical world structure and camera trajectory.
We evaluate our approach on several sequences from the public KITTI dataset and demonstrate quantitatively that it improves camera and object tracking compared to state-of-the-art approaches.
}
\end{abstract}
~\\
\begin{keywords}
SLAM, Localization, Mapping
\end{keywords}
\section{INTRODUCTION}

\label{sec:intro}

Visual Simultaneous Localization And Mapping (SLAM) is an important problem for robotics that has been heavily studied in the past decade. Its goal is to estimate the pose of a camera moving in a scene while building a map of the environment. Some algorithms such as \cite{mur2017orb, engel2014lsd} can efficiently solve this problem, however they rely on the static scene assumption. This hypothesis assumes that the world is a single rigid body and thus that no object can move within it. \rev{This assumption, which is rarely met in most real world scenes, as they can contain moving objects (e.g. cars, bikes, pedestrians for urban scenes of the KITTI dataset)}, limits the scenarios in which a SLAM algorithm can be used. Classical SLAM systems such as \cite{mur2017orb} try to alleviate this assumption using robust estimators, allowing them to flag moving parts as outliers. However as soon as the number of moving points is too important, the estimated camera pose accuracy decreases. This makes this approach unsuitable for some scenes (e.g. crowded or urban scenes). Some systems \cite{bescos2018dynaslam, yu2018ds} have been proposed to detect and mask out dynamic objects in images, thus making the static scene assumption valid. However some recent approaches \cite{bescos2021dynaslam, zhang2020vdo, runz2018maskfusion, huang2020clustervo} argue that moving objects represent valuable information that can be necessary for some applications. 
Most recent approaches trying to solve both SLAM and object tracking have used semantics as an additional source of information. Semantic knowledge can indeed be beneficial to SLAM \cite{cadena2016past, rosinol2020kimera} as it contains information about the class dynamicity \cite{bescos2018dynaslam} which is higher level information than simple 3D points. 
\begin{figure}
    \centering
    \includegraphics[width=\columnwidth]{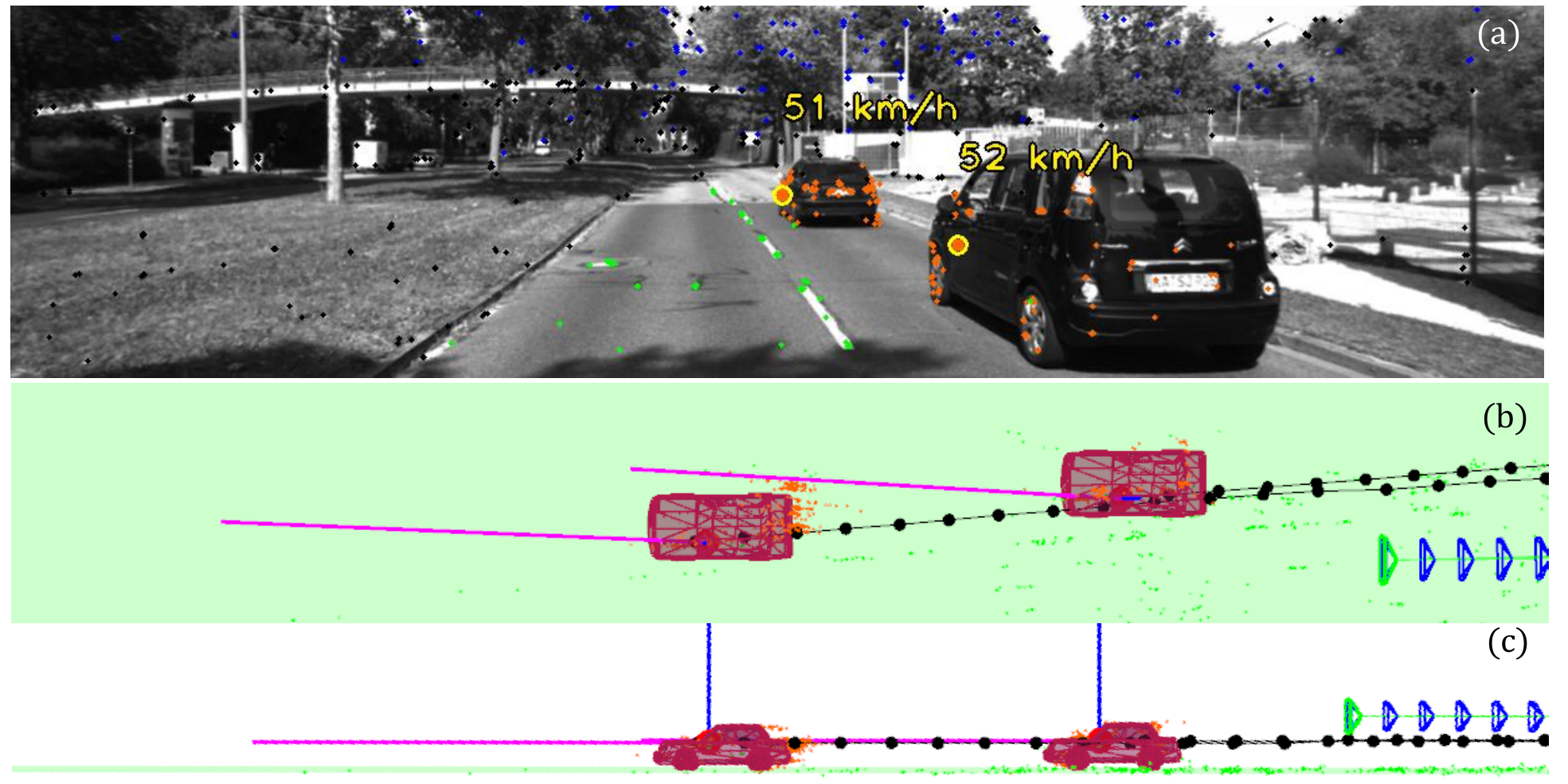}
    \caption{Our approach allows us to track objects in the scene such as cars. Here we can see: (a) the frame with semantic points and tracked clusters (orange cars) with their estimated speed. (b) a map top view with tracked clusters (orange cars), clusters trajectories (black spheres and lines), clusters twists (blue and purple lines), road points and plane (in green) and camera trajectory (green and blue frustums). (c) Map side view. The rotation part of the twists (blue lines) is perpendicular to the road
plane, the translation part (purple lines) is parallel to the plane.}

\vspace{-5mm}
    
    \label{fig:first_fig}
\end{figure}


In this paper we present a stereo SLAM system called TwistSLAM, as we estimate objects' twists to track them and consider that objects are linked to each other through mechanical joints, similarly to joints linking different parts of a robot. 
An illustration of our algorithm is visible in figure \ref{fig:first_fig}: the camera pose is estimated simultaneously with all moving objects in the scene and the map structure (here the plane of the road) constrains the movement of objects.
Our approach is based on ORB-SLAM2 \cite{mur2017orb} and S$^3$LAM \cite{gonzalez2021s3lam}. In our work we use semantic information to build a map of clusters corresponding to objects in the scene. The clustering of the scene allows us to estimate the pose of the camera using static clusters only such as  \textit{road} or \textit{house}. The other clusters that can be dynamic are tracked and their pose is updated in the map through the estimation of twists that represent their velocity. 
Most SLAM systems that can track dynamic objects directly estimate their pose through the minimization of a reprojection error function \cite{bescos2021dynaslam, huang2020clustervo} or with 3D points registration \cite{runz2018maskfusion}. Doing so, the estimated pose of an object has 6 degrees of freedom. This does not correspond to reality. For example a car has only 3 degrees of freedom, 2 translations in the road plane and 1 rotation around its normal, hence its pose should be constrained. Our goal is thus to remove degrees of freedom corresponding to physically unfeasible movements.
To do so we choose to represent those constraints as mechanical joints which makes our approach highly generic. A mechanical joint between clusters constrains the estimated twist of a dynamic cluster by blocking some of its degrees of freedom thus reducing the effect of noise on the estimation. Once an object twist has been estimated it can be used to update the object pose which enables object tracking. 
The object poses can then be tightly refined with camera poses and 3D points within a bundle adjustment that also applies mechanical joints constraints.

~\
Our contributions presented in this paper are:
\begin{itemize}

\item A semantic SLAM system that can robustly estimate the pose of a camera in static as well as dynamic scenes.
\item A stereo SLAM framework that can track multiple moving objects in the scene.
    
    \item A new formulation for both the tracking and bundle adjustment that takes into account the characteristics of mechanical joints between objects in the scene \rev{to constrain their movements}.

\end{itemize}
\rev{We also evaluate our approach on several sequences from the public KITTI dataset and compare our approach quantitatively with respect to ORB-SLAM2 \cite{mur2017orb}, DynaSLAM \cite{bescos2018dynaslam}, VDO-SLAM \cite{zhang2020vdo} and DynaSLAM II \cite{bescos2021dynaslam}, which demonstrates the benefits of our method in terms of object poses and velocities and camera pose estimation accuracy.}
The rest of the paper is described as follows. First we describe related work on  dynamic classical and semantic SLAM. Then we give an overview of mathematical concepts that will be used in this paper.
Following, we describe our approach to build a semantic map of clusters, estimate the pose of a camera in a dynamic scene, track moving clusters within the scene while using joint constraints to improve object tracking and refine all estimations with a bundle adjustment. Finally we demonstrate the benefits of our approach on multiple sequences from a public dataset.
\section{\rev{Related work: Dynamic SLAM}}
\rev{In this section we first present some classical \cite{mur2017orb} and dynamic SLAM systems \cite{li2017rgb} that remove dynamic outliers. Then we focus on semantic SLAM systems that tackle the problem of dynamic objects by masking out \cite{bescos2018dynaslam, yu2018ds} or tracking \cite{runz2018maskfusion, bescos2021dynaslam, huang2020clustervo, zhang2020vdo, yang2019cubeslam} objects in the scene. Our approach belongs to this last class of systems.} For a more in-depth survey of SLAM we refer the reader to \cite{cadena2016past, taketomi2017visual}.

ORB-SLAM2 \cite{mur2017orb}, which follows the work of \cite{mouragnon2006real, klein2007parallel} uses three parallel threads to estimate the pose of a camera through relocalization or tracking, build a map, refine both using a bundle adjustment \cite{triggs1999bundle} and close looping trajectories. ORB-SLAM2 manages to mitigate the influence of dynamic objects through the use of a RANSAC scheme and a robust cost function based estimation.
To tackle this problem some methods propose to roughly estimate the pose of the camera and robustly find outliers in the scene or in the image. Outliers are then removed or downweighted and the camera pose is refined. For example \cite{li2017rgb} proposes a direct approach based on the alignment of depth edges using an ICP scheme. For each point they robustly estimate a staticity confidence score which downweights dynamic objects and an intensity assisted ICP robsulty refines the pose using those weights. 
\rev{However those approaches fail when the amount of dynamic parts in the image is too high and are not able to track dynamic objects.}
DS-SLAM \cite{yu2018ds} applies a geometrical moving consistency check on segmented areas. This score allows them to know which areas correspond to moving objects, which are then discarded for robust camera pose estimation and mapping. DynaSLAM \cite{bescos2018dynaslam} uses semantic information to segment a priori moving objects, which are not used for tracking and mapping. The segmentation is refined using depth information. This approach improves camera localization in dynamic scenes but deteriorates it when a priori moving objects are in reality static such as parked cars. 
\rev{Those approaches, similarly to the previous ones, are not able to track dynamic objects.}
MaskFusion \cite{runz2018maskfusion} is one of the first semantic dynamic SLAM that can track objects. Inspired from \cite{runz2017co} it makes use of 2D instance segmentation to detect objects in the scene and tracks them using both photometric and geometric information from an RGB-D camera.
DynaSLAM II \cite{bescos2021dynaslam} uses semantic information to detect objects. Object 3D points are represented in the object reference frame and used to estimate the object pose at all time by minimizing their reprojection error. ClusterVO \cite{huang2020clustervo} is similar to \cite{bescos2021dynaslam}, but they consider object detection (i.e. 2D bounding boxes) as input which is much faster to infer than dense masks. They also apply a cleaning procedure to improve dynamic keypoints matching and make sure that 2D points do not come from the background of the bounding box. 
VDO-SLAM \cite{zhang2020vdo} proposes to use optical flow to track features extracted more densely than other systems, which allows them to obtain a more precise object pose estimation. Furthermore the optical flow and the object and camera motions are tightly refined. CubeSLAM \cite{yang2019cubeslam} is different from the previously cited papers as it is an object based SLAM. A 3D bounding box is fitted for each object detected by a CNN (such as \cite{redmon2016you}), which allows them to know the object 6 DoF pose and dimensions for each image. The bounding boxes poses are then optimized along with the camera poses in a single BA, similarly to \cite{bescos2021dynaslam}.  

\section{Background on twists and homogeneous transformations}
In this section we present the mathematical concepts and notations that we use in this paper. 
\rev{A rigid object $o$, which can move in the 3D scene, can be associated with its coordinate frame $\mathcal{F}_{o}$. Its pose relative to the world coordinate frame $\mathcal{F}_{w}$ can be represented by the homogeneous matrix in the $i^{th}$ frame}
\begin{equation}
    \objposetime{i} = \begin{pmatrix}
    \rot{w}{o_i} & \trans{w}{o_i} \\
    0 & 1 \\
    \end{pmatrix} \quad \in \quad \mathrm{SE}(3)
\end{equation}
where $\mathrm{SE}(3)$ denotes the special euclidean group. \rev{This matrix maps points expressed in the object coordinate frame and denoted $\point{o}$ to points expressed in the world coordinate frame and denoted $\point{w}$,  according to the following equation: $
    \point{w} =~\objpose \point{o}$.}
The velocity of a moving object can be represented using a twist $\twist$ defined as 
\begin{equation}
    \twist = 
    \begin{pmatrix}
    v_x & v_y & v_z & \omega_x & \omega_y &    \omega_z 
    \end{pmatrix}^\top = \begin{pmatrix}
    \tspeed &
    \rspeed 
    \end{pmatrix}^\top \in \mathbb{R}^6
\end{equation}
where the first 3 components $\tspeed = (v_x, v_y, v_z)^\top \in \mathbb{R}^3$ denote the translational velocity and the other components $\rspeed = (\omega_x, \omega_y, \omega_z)^\top \in \mathbb{R}^3$ represent the rotational velocity. There exists a matrix representation for twists that can be obtained using the operator $[.]_{\wedge}$ defined by:
\begin{equation}
    [\twist]_{\wedge} = 
    \begin{pmatrix}
    [\rspeed]_{\times} & \tspeed \\
    0 & 0 \\
    \end{pmatrix} \in \mathrm{se}(3)
\end{equation}
\rev{
where $[.]_{\times}$ is the skew-symmetric operator defined such as for $\bf{a}$ $= (a_x, a_y, a_z)^\top$ $\in \mathbb{R}^3$:
\begin{equation}
    [\bf a]_{\times} = \begin{pmatrix}
    0 & -a_z &a_y \\
    a_z & 0& -a_x \\
    -a_y & a_x & 0 \\
    \end{pmatrix}
\end{equation}
}
and $\mathrm{se}(3)$ is the Lie algebra associated to $\mathrm{SE}(3)$. We denote $\twistcoord{o_i}{w}$ the twist corresponding to the velocity of the object $o$ at frame $i$ expressed in the world coordinate frame. Similarly, $\twistcoord{o_i}{o_i}$ is the velocity of the object $o$ at frame $i$ expressed in its own coordinate frame.
As velocities can be integrated over time to obtain new positions, there exists a mapping from $\mathrm{se}(3)$ to $\mathrm{SE}(3)$ called the exponential map and denoted:
\begin{equation}
    \exp: \, [\twist\delta t]_\wedge \in \mathrm{se}(3) \, \longrightarrow \, \mathbf{T} \in \mathrm{SE}(3)
\end{equation}
where $\delta t$ is the time interval duration \cite{blanco2010tutorial}.

Using the exponential map we can recover the pose of the object $o$ moving according to the twist $\twistcoord{o_i}{w}$ from its initial pose at frame $i$, $\objposetime{i}$ to its next pose at frame $i+1$, $\objposetime{i+1}$ using the following formula:
\begin{equation}
    \objposetime{i+1} = \exp(\twistcoord{o_i}{w}\delta t_i) \objposetime{i} = \objposetime{i} \exp(\twistcoord{o_i}{o_i}\delta t_i)
    \label{eq:expmap}
\end{equation}
Note that the choice of coordinate frame matters, it can be useful to define an operator to change the coordinate frame of a twist. Such operator is called the adjoint map $\adj{w}{o_i} \in \mathbb{R}^{6\times6}$ and links twists in different coordinate frames according to:
\begin{equation}
    \twistcoord{o_i}{w} = \adj{w}{o_i} \; \twistcoord{o_i}{o_i}
\end{equation}
The adjoint map can be computed using the relative pose $\pose{w}{o_i}$ between $\coordframe{o_i}$ and $\coordframe{w}$:
\begin{equation}
    \adj{w}{o_i} = 
    \begin{pmatrix}
    \rot{w}{o_i} & [\trans{w}{o_i}]_\times \; \rot{w}{o_i} \\
    0 &\rot{w}{o_i} \\
    \end{pmatrix}
    \label{eq:adjoint}
\end{equation}
For simplicity we will consider in the remainder of this paper that $\delta t = 1$ without loss of generality.
\section{TwistSLAM: Constrained SLAM in Dynamic Environment}
\begin{figure}
    \centering
    \includegraphics[scale=0.4]{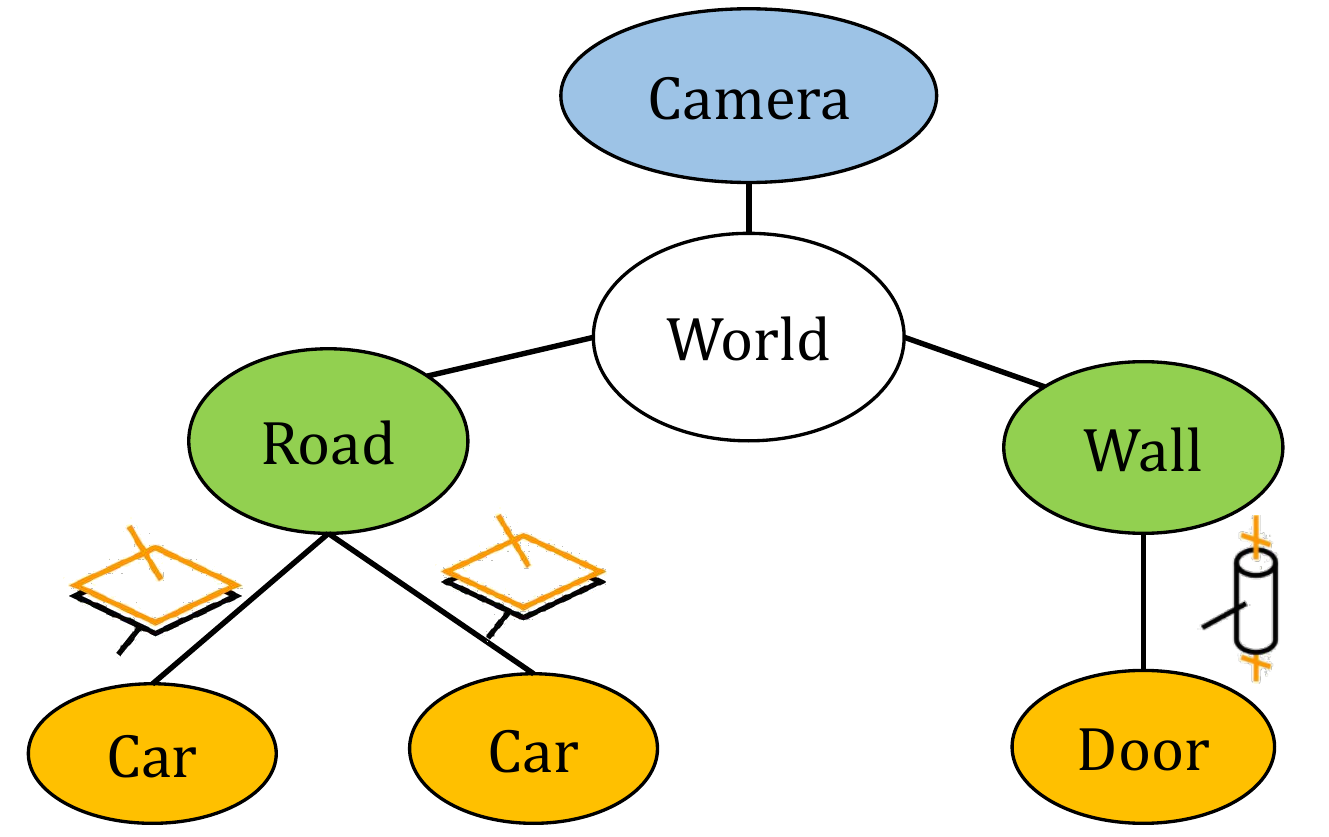}
    \caption{Example of semantic graph: dynamic clusters are linked to static parent clusters with mechanical joints such as \textit{planar} or \textit{revolute}.}
    
    \label{fig:graph}
    
\vspace{-5mm}
\end{figure}

In this section we present our approach. 
Our general idea is to represent the world as a graph of semantic clusters, which is similar to a scene graph and can be seen in figure \ref{fig:graph}. The vertices of the graph correspond to objects in the scene and the edges to physical links that exist between objects. Our goal is to estimate the pose of the camera and the pose of every moving object while using mechanical joints between objects to improve those estimations.
For example both clusters \textit{car} in our graph are linked to the road with a planar constraint that allows only 3 degrees of freedom: a rotation around the normal of the plane and 2 translations within the plane. Such simple representations allow us to be highly generic as, for a given semantic class, we only need to define its static parent and the type of mechanical joint.

\begin{figure*}
    \centering
    \includegraphics[width = 0.86\textwidth]{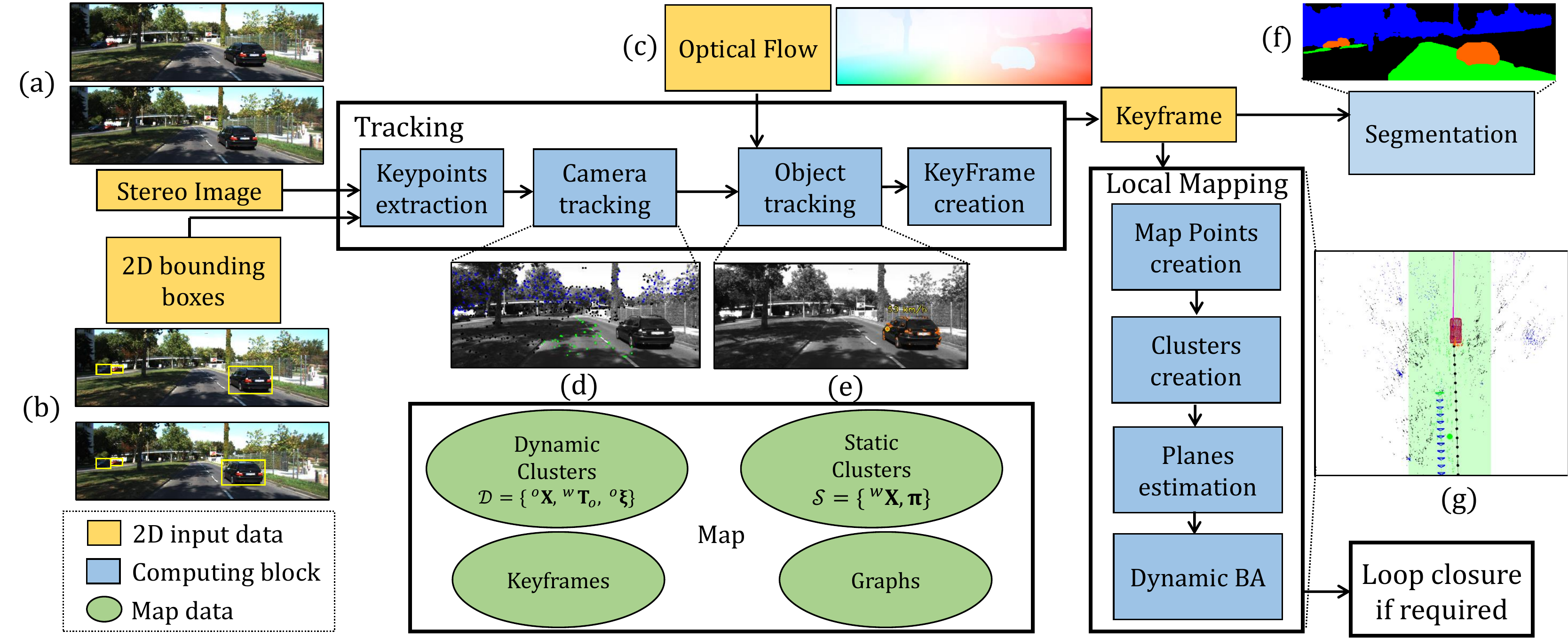}
    \caption{The pipeline of our approach: static keypoints are extracted from a stereo image (a) and used for camera tracking (d), dynamic keypoints are extracted from bounding boxes within the stereo images (b) and matched using optical flow (c) with the previous frame to track dynamic objects (e). The keyframe is then segmented (f) to create new semantic map points and clusters. Finally the object and camera poses are jointly refined with the dynamic and static map points in a BA (g). }
    \vspace{-5mm}
    \label{fig:main_fig}
\end{figure*}

The pipeline of our approach is presented in figure \ref{fig:main_fig}. Using semantic information we create clusters of points corresponding to objects in the scene. Then, we use static semantic clusters (e.g. road, floor, house) to robustly track the camera, even in dynamic scenes. Next, we match keypoints corresponding to dynamic objects (e.g., car, bike,...) to either track them or triangulate new 3D points using stereo information. All poses estimations from the camera and the objects are then refined with static and dynamic 3D points in a bundle adjustment process. 

The main novelty of our approach comes from the fact that we optimize the velocities of dynamic objects rather than their pose and constrain the velocities according to mechanical joints between objects. This approach is highly generic as we only need to define a handful of joints (that correspond to normalized joints in mechanics) and a list of semantic classes pairs for each joint (e.g. the wall-door joint corresponds to a revolute joint, the car-road joint corresponds to a planar joint). As we will latter show, it allows us to remove displacements along directions that are not physically possible (e.g. a car translating vertically). This allows us to obtain a more precise estimation of the dynamic object poses. 
 
\subsection{Creating Clusters from panoptic segmentation}
Most recent semantic dynamic SLAM systems use either an object detection  or an instance segmentation algorithm. Working in the continuity of S$^3$LAM \cite{gonzalez2021s3lam} we chose to estimate the panoptic segmentation of images, obtained using \cite{wu2019detectron2}. This allows us to know the semantic class of each pixel in the image and to give a unique id to each object.  
Similarly to \cite{gonzalez2021s3lam} we fuse multiple 2D observations of a single 3D point to obtain its class and id. Doing so we obtain a semantic map, which allows us to create a set of $K$ clusters $\mathcal{O} = \{O_{k}, k \in [1,K]\}$. A cluster is a set of 3D points corresponding to a single object in the scene. Points are grouped according to their class and instance id. The set of clusters can be expressed as the set of a priori static clusters $\mathcal{S}$ (such as \textit{road, building, ...}) and the set of a priori dynamic clusters $\mathcal{D}$ (such as \textit{car, bike, human, bus, ...}).
As static clusters are fixed, we represent their 3D points $\{\point{w}\}$ in the world frame. In contrast, each dynamic cluster contains a set of 3D points $\{\point{o}\}$ expressed in the object coordinate frame, a set of poses $\{\pose{o}{w}\}$ and a set of twists $\{\twistcoord{o}{w}\}$ representing the cluster trajectory and velocity through time. For simplicity in the remainder of this paper we will omit the object index $k$ as its use is straightforward.

\subsection{Clusters geometry}
Our goal is to constrain the velocity of moving clusters according to mechanical joints. To do so we need to estimate the pose of those joints. We propose to do this using the estimated geometry of some clusters. We chose to consider only planar clusters, which allows our approach to be highly generic as planes are common in man-made environment.
For clusters corresponding to a priori chosen classes (such as the road or the facade of a building) we estimate a 3D plane, represented by $\pi = (a,b,c,d)^\top$ with $||\pi||^2=1$, using only its 3D points $\{\point{w}\}$. The plane follows the following equation: $\pi^{\top} \point{w} = 0$ and can be estimated using an SVD. To make it robust to outliers we use a RANSAC scheme.

\subsection{Dynamic SLAM}
As we do not know which dynamic objects in the scene are really moving we chose to estimate the camera pose using only static objects. Using points from static clusters we minimize the following cost function \begin{equation}
    E(\camposetime{i}) = \sum_{j \in \mathcal{S}}{\rho(||\imgpointij{i}{j}-\proj(\camposetime{i}, \pointidx{w}{j})||_{\Sigma_{i,j}^{-1}})} 
    \label{eq:track_stat}
\end{equation}
where $\imgpointij{i}{j}$ is the 2D keypoint corresponding to the observation of $\pointidx{w}{j}$ in the $i^{th}$ frame, $\proj$ is the pinhole camera projection function, $\rho$ is a robust cost function (in our case Huber) \cite{malis2006experiments} and $\Sigma_{i,j}$ is the covariance matrix of the reprojection error. Doing so the estimated camera pose does not take into account potentially moving objects, hence it is robust in dynamic scenes. However the estimation can be deteriorated in scenes that contain many potentially moving objects that are in reality static, like for example parked cars. To solve this problem, we chose to estimate the pose of all moving objects and integrate them in the bundle adjustment, so that the velocity of static objects is close to 0 and their points act as static points. 
\subsection{Dynamic data association and keypoints} \label{sec:data_association}
Dynamic data association is a challenging problem for two reasons: first the combination of the camera and the object movements can produce large displacements in the image space thus needing a large radius search for keypoints matching. Second a large movement can cause an important visual variation of the object in the image (e.g. due to luminosity changes on the object or to viewpoint changes) which makes the matching process more difficult. To overcome those challenges we propose to use the optical flow estimation produced by a CNN \cite{teed2020raft} to have a good estimate of the keypoints location and reduce the search radius, thus reducing both search time and the probability of false matches.

One problem of object tracking compared to classical SLAM is that dynamic objects usually occupy a small portion of the image, leading to too few object points to obtain a precise estimation \cite{bescos2021dynaslam, zhang2020vdo}. To solve this problem we force the keypoint extraction process to keep more keypoints from areas defined by objects bounding boxes. 
The keypoints are then used either to create new 3D points with stereo triangulation, which are added to existing clusters or used to create new clusters, or used to track the existing cluster.
\subsection{Mechanical joints as inter-cluster constraints} 
Using matches found by the data association process we seek to estimate the pose of dynamic objects in the scene.
Our assumption in this work is that many moving clusters can be represented as being linked to a static parent cluster with a specific mechanical joint. There exist 12 normalized joints (ISO 3952) that can be associated with the degrees of freedom they have. For example the planar joint has 3 degrees of freedom: 2 translations in the plane and 1 rotation around its normal, this joint can represent the displacement of a car relative to its static parent, the road. Another example is the revolute joint which has a single degree of freedom corresponding to the rotation around a single axis. In this case the static parent cluster is the wall, the moving cluster is the door and its only possible movements are rotations around the axis of the joint (corresponding to the hinge).

To easily model all types of joints, similarly to \cite{comport2007kinematic}, we propose to decompose the space of twist as the sum of two orthogonal spaces:
\begin{equation}
    \mathbb{R}^6 = \mathbb{F}_l + \mathbb{F}_l^{\top}
\end{equation}
where $\mathbb{F}_l$ (which stands for freedom) is the space of twists allowed by the mechanical joint $l$ with coordinate frame $\mathcal{F}_l$.  In the case of a planar joint with axis $z$, $\mathbb{F}_l$ is defined as:
\begin{equation}
    \mathbb{F}_l = Span(
    \begin{pmatrix}
    1 & 0 & 0&0&0&0 \\
    0 & 1 & 0&0&0&0 \\
    0 & 0 & 0&0&0&1
    \end{pmatrix}^\top)
\end{equation}
where $Span$ is the linear span \cite{axler2014linear}. In general we note:
\begin{equation}
    \mathbb{F}_l = Span(\mathbf{A}_l)
\end{equation}
where $\mathbf{A}_l$ is a basis of $\mathbb{F}_l$.
To make the displacement of an object physically accurate, its twists have to lie within the $\mathbb{F}_l$ space. 
To do so we project the twist from its original space to $\mathbb{F}_l$. This is straightforward as $\mathbb{R}^6$ is Euclidean, the operation projector is a $6\times6$ matrix defined as:
\begin{equation}
    \bm{\Pi}_l = \mathbf{A}_l(\mathbf{A}_l^{\top}\mathbf{A}_l)^{-1}\mathbf{A}_l^{\top}
\end{equation}

In the example of a planar joint, it is easy to compute that:
\begin{equation}
    \bm{\Pi}_l  = \begin{pmatrix}
    1 & 0& 0 & 0 & 0 & 0 \\
    0 & 1& 0 & 0 & 0 & 0 \\
    0 & 0& 0 & 0 & 0 & 0 \\
    0 & 0& 0 & 0 & 0 & 0 \\
    0 & 0& 0 & 0 & 0 & 0 \\
    0 & 0& 0 & 0 & 0 & 1 \\
    \end{pmatrix}
\end{equation}
in that case, a general twist can be projected such that:
\begin{equation}
    \bm{\Pi}_l \; \twist = \begin{pmatrix}
    v_x & v_y & 0 & 0 & 0 &\omega_z
    \end{pmatrix}^\top
\end{equation}
As we can see the only remaining degrees of freedom of the twist are coherent with the joint. Using this new constraint, we can modify the reprojection equation:
\begin{equation}
    \imgpointij{i}{j} = \proj(\camposetime{i} \exp{(\bm{\Pi}_l \twistcoord{o_i}{w})\objposetime{i-1}, \pointidx{o}{j}})
\end{equation}
However this equation is only true if the twist is expressed in the joint coordinate frame, yet according to (\ref{eq:expmap}), it is naturally expressed either in the world or in the object coordinate frame. To change the coordinate frame of a twist we can use the adjoint map defined in (\ref{eq:adjoint}). Hence the reprojection equation of the $j^{th}$ point in frame $i$ becomes:
\begin{equation}
    \imgpointij{i}{j} = \proj(\camposetime{i} \exp{(\adj{w}{l}\; \bm{\Pi}_l \; \adj{l}{w} \; \twistcoord{o_i}{w})\objposetime{i-1}, \pointidx{o}{j}})
\end{equation}
In the remainder of this paper we will note $\projmat = \adj{w}{l}\; \bm{\Pi}_l \; \adj{l}{w}$ for simplicity. This equation takes a 3D point in the object frame, transforms it in the world frame using the previous object pose and multiplies it by the exponential of the current twist to get its current position. The twist is expressed in the joint coordinate frame with the adjoint map, projected using $\bm{\Pi}_l$ to keep only the relevant components and expressed again in the world frame with the inverse adjoint map. Doing this we obtain a 3D point in the world frame for frame $i$ with a transformation that perfectly respects the mechanical joint. We then apply the camera pose to obtain the point in the camera coordinate frame, which allows us to project it in the image. 

~\\
Using the reprojection function we can estimate the twist corresponding to the transformation of a set of object points between frame $i-1$ and $i$ by minimizing the following error:
\begin{equation}
\medmuskip=-2mu
\thinmuskip=-2mu
\thickmuskip=-1mu
    E(\twistcoord{w}{o_i}) =  \sum_j{\rho(||\imgpointij{i}{j}-\proj(\camposetime{i} \exp{(\projmat \twistcoord{o_i}{w})\objposetime{i-1}, \pointidx{o}{j}})||_{\Sigma_{i,j}^{-1}})}
    \label{eq:dyna_tracking}
\end{equation}
where $\rho$ is the Huber robust estimator \cite{malis2006experiments} and $\Sigma_{i,j}$ is the covariance matrix of the reprojection error. In \cite{mur2017orb} the convariance matrix depends on the scale at which the keypoints are observed. In our case we choose to estimate it using the median absolute deviation (MAD) \cite{malis2006experiments} that is a robust estimator of the standard deviation of the reprojection error.
We perform the optimization using the Levenberg-Marquardt algorithm on matches found between the current and the previous frame. Then we refine this twist with an approach similar to \cite{mur2017orb} by projecting map points, transformed with the estimated twist, in the current frame to search for additional matches and obtain a more accurate estimation.
The object pose in frame $i$ is then updated as $\objposetime{i} = \exp(\projmat \twistcoord{o_i}{w})\objposetime{i-1}$. This tracking procedure is repeated for all objects. 

\subsection{Dynamic Bundle Adjustment}
The goal of classical bundle adjustment is to refine the camera trajectory and 3D points position estimation. 
The dynamic bundle adjustment has multiple goals. First, the refinement of the dynamic objects trajectory and their 3D points position, jointly with the camera trajectory and 3D static points position. Second, it allows to link the object and the camera trajectory, indeed if the bundle adjustment did not take into account dynamic objects, only the camera pose would have an impact on the object pose, which would not improve it. By taking into account dynamic points whose position is estimated over time we can use them to refine the camera pose, similarly to static points but with less accuracy since object pose estimation is more noisy. Finally, it allows us to apply a soft constrain on twists within a temporal window. Doing so we obtain smoother trajectories and velocities that are more physically plausible.

Our bundle adjustment cost function can be written as follows:
\begin{equation}
    E(\{\tildetwistcoord{o}{w}, \campose, \point{w},  \point{o}\}) = \sum_{i,j} e_{stat}^{i,j}+\sum_{i,j}e_{dyna}^{i,j}+\sum_{i}e_{const}^{i} \label{eq:ba}
\end{equation}
where $e_{stat}^{i,j}$ is the classical static reprojection error:
\begin{equation}
\nonumber
    e_{stat}^{i,j} = \rho(||\imgpointij{i}{j} - \proj(\camposetime{i}, \pointidx{w}{j})||_{\Sigma_{i,j}^{-1}})
\end{equation}
$e_{dyna}^{i,j}$ is a dynamic reprojection error: 
\begin{equation}
\nonumber
   e_{dyna}^{i,j} = \rho(||\imgpointij{i}{j} - \proj(\campose \exp(\projmat\tildetwistcoord{o_i}{w})\objposetime{i}, \point{o})||_{\Sigma_{i,j}^{-1}})
\end{equation}
where $\Sigma_{i,j}^{-1}$ is estimated using the MAD as in equation (\ref{eq:dyna_tracking}).
And $e_{const}^{i}$ is a constant velocity model that penalizes twists variations by linking 3 consecutive poses:
\begin{equation}
\nonumber
    e_{const}^{i} = \rho(||\projmat\twistcoord{o_{i+1}}{w}-\projmat\twistcoord{o_i}{w}||_{\mathbf{W}})
\end{equation}
where $\mathbf{W}$ is a diagonal weight matrix used to balance the errors,  $\twistcoord{o_{i+1}}{w}$ is the twist linking the poses $\exp(\projmat\tildetwistcoord{o_i}{w})\objposetime{i}$ and $\exp(\projmat\tildetwistcoord{o_{i+1}}{w})\objposetime{i+1}$ and $\twistcoord{o_{i}}{w}$ is the twist linking the poses $\exp(\projmat\tildetwistcoord{o_{i-1}}{w})\objposetime{i-1}$ and $\exp(\projmat\tildetwistcoord{o_{i}}{w})\objposetime{i}$. Those twists are computed using the logmap from $\mathrm{SE}(3)$ to $\mathrm{se}(3)$ defined in \cite{blanco2010tutorial} and can be written for $\twistcoord{o_{i+1}}{w}$ as:
\begin{equation}
\nonumber
    \tildetwistcoord{o_{i+1}}{w} = \log(\exp((\projmat\tildetwistcoord{o_{i+1}}{w})\objposetime{i+1})(\exp(\projmat\tildetwistcoord{o_i}{w})\objposetime{i})^{-1})
\end{equation}

~\\
This equation moves each pose while respecting the mechanical joints constraints.
Optimizing it can be cumbersome however the Schur trick can be applied as its Hessian is sparse \cite{bescos2021dynaslam}. 
These equations are classically optimized on a set of local keyframes that share visual information, but in our case, inspired by \cite{huang2020clustervo} we chose to have 2 sets of keyframes: temporal and spatial. All frames are converted to temporal keyframes to improve the tracking of fast moving objects and be able to track an object as soon as it enters the field of view of the camera. Keyframes stay in the temporal set for a fixed duration (in our case 5 seconds) they are then culled more severely than in ORB-SLAM2. This allows us to keep a reasonable number of keyframes. We chose to optimize camera poses on the set of temporal and local keyframes, while object poses are only optimized on the set of temporal keyframes and fixed in all other keyframes. Doing so, we apply our constant motion model only on the temporal window, allowing clusters to accelerate or decelerate. 

\subsection{Computing the cost functions jacobians}
To optimize the cost functions (\ref{eq:dyna_tracking}) and (\ref{eq:ba}) with a Levenberg-Marquardt optimizer we need to compute their jacobian. 
First we compute the Jacobian of the cost function used for object tracking $E(\twistcoord{w}{o_i})$. Using the chain rule and getting inspiration from \cite{blanco2010tutorial} it can be shown that the Jacobian for the $j^{th}$ point is:
\begin{gather}
\begin{aligned}
&(J_E)_j = 
    (\frac{\partial E(\twistcoord{w}{o_i})}{\partial\twistcoord{w}{o_i}})_j  \\
    &= \frac{\partial \proj(\point{c}_j)}{\partial \point{c}_j}(\point{c}^{\top}_j \otimes \mathbf{I}_3)(\mathbf{I}_4 \otimes \rot{c_i}{w})(\objposetime{i-1}^\top \otimes \mathbf{I}_3) \partial_{\exp}\projmat
    \raisetag{45pt}
\end{aligned}
\end{gather}
where $\otimes$ is the Kronecker product, $\mathbf{I}_N$ is an identity matrix of size $N$, $\rot{c_i}{w}$ is the rotation matrix of $\camposetime{i}$ and:
\begin{equation}
\partial_{\exp} = 
\frac{\partial \exp(\twist) }{\partial \twist} = 
    \begin{pmatrix}
    \mathbf{0}_{3\times 3} & -[\mathbf{e}_1]_{\times} \\
    \mathbf{0}_{3\times 3} & -[\mathbf{e}_2]_{\times} \\
    \mathbf{0}_{3\times 3} & -[\mathbf{e}_3]_{\times} \\
    \mathbf{I}_{3} & \mathbf{0}_{3\times 3} \\
    \end{pmatrix}
\end{equation}
where $\{\mathbf{e}_1, \mathbf{e}_2, \mathbf{e}_3\}$ is the canonical base of $\mathbb{R}^3$.
 Then, we compute the Jacobian of $e_{dyna}$ in equation (\ref{eq:ba}), which is very similar to the previous jacobian. The derivatives of the function with respect to camera poses and points are the same as for classical bundle adjustment \cite{dellaert2014visual}.
 For the object poses we can compute:
\begin{gather}
\begin{aligned}
\medmuskip=-2mu
\thinmuskip=-2mu
\thickmuskip=-2mu
   & (J_{e_{dyna}})_{i,j} = \frac{\partial e_{dyna}^{i,j}}{\partial\tildetwistcoord{w}{o_i}} \\=   & \frac{\partial \proj(\point{c}_j)}{\partial \point{c}_j}(\point{c}^{\top}_j \otimes \mathbf{I}_3)(\mathbf{I}_4 \otimes \rot{c_i}{w})(\objposetime{i}^\top \otimes \mathbf{I}_3) \partial_{\exp} \projmat
    \raisetag{45pt}
\end{aligned}
\end{gather}

Finally we compute the jacobian of the constant velocity constrain with respect to each of the 3 twists involved in the constrain:
\begin{gather}
\begin{aligned}
    J_{e_{const}} = 
    \begin{pmatrix}
    \frac{\partial e_{const}}{\partial\tildetwistcoord{o_{i-1}}{w}} &  
    \frac{\partial e_{const}}{\partial\tildetwistcoord{o_i}{w}} & 
    \frac{\partial e_{const}}{\partial\tildetwistcoord{o_{i+1}}{w}} \\
    \end{pmatrix}
\end{aligned}
\end{gather}
we only show here the left part of the jacobian as the other parts are similar.
\begin{gather}
    \begin{aligned}
        \frac{\partial e_{const}}{\partial\tildetwistcoord{o_{i-1}}{w}} = \projmat  \frac{\partial \log(\mathbf{T})}{\partial \mathbf{T}}(\mathbf{I}_4 \otimes \mathbf{R}) \partial_{\exp}\projmat
    \end{aligned}
\end{gather}
with $\mathbf{T}$ = $\exp((\projmat\tildetwistcoord{o_{i}}{w})\objposetime{i})(\exp(\projmat\tildetwistcoord{o_{i-1}}{w})\objposetime{i-1})^{-1}$, $\mathbf{R}$ is the rotation matrix of $\exp((\projmat\tildetwistcoord{o_{i}}{w})\objposetime{i})(\objposetime{i-1})^{-1}$ and the derivative of the logmap is given by \cite{blanco2010tutorial}.


\section{Experiments}
In this section we present the experiments we conducted to test our approach. We evaluate both the accuracy of the camera pose estimation and of the object pose estimation. 
\subsection{Experiments details}
\textbf{Datasets.}
We evaluate our approach on the KITTI \cite{geiger2012we} tracking dataset which consists in multiple sequences recorded from a camera mounted on a car. This dataset is particularly interesting for our approach as it contains the ground truth for both the camera pose and for some objects poses such as vehicles. 
\rev{It should be noted that the segmentation network can yield an "unknown" class which we consider to be static, as the dynamic classes in the KITTI dataset (cars, bikes and pedestrians) are correctly segmented by the network.}
~\\
\textbf{Metrics.}
The metrics for the evaluation of SLAM systems are usually the absolute translation error (ATE) \cite{sturm12iros} and the relative pose error (RPE) \cite{zhang2018tutorial}. For each sequence we report the translation and rotation parts of the RPE, as it is done by both VDO-SLAM and DynaSLAM2.  
The object pose estimation accuracy can be evaluated using 2 different types of metrics: on the one hand the ATE and RPE that measure the quality of the objects trajectories and on the other hand the MOTP that evaluates the per-frame accuracy of objects 3D bounding boxes estimations and that we compute similarly to \cite{bescos2021dynaslam} using KITTI evaluation tools. As we do not estimate object boxes we use the ground truth box at the first pose of each object and propagate it using our camera and object pose estimations. 
We evaluate the true positive rate (TP) and the MOTP using the projected 3D bounding box (2D), in bird view (BV) and in 3D.
~\\

\subsection{Camera pose estimation}
In this subsection we evaluate the accuracy of our camera pose estimation. Similarly to \cite{bescos2021dynaslam} we only show here sequences in which the camera is moving.
As we can see in table \ref{table:ego} our approach improves camera pose estimation on several sequences. 
\rev{To evaluate the stability of our approach we also computed the standard deviation over 10 runs for the sequence 3 and obtained a value of 1 $\times 10^{-4}$ and $4 \times 10^{-4}$ for the translation and the rotation respectively.}
As objects are often either small, only visible for a short time or static, \cite{mur2017orb} performs well, but as we track clusters using many points, with a good precision, especially for clusters that do not move, we are able to reduce the drift.
\cite{zhang2020vdo} also gives good results but requires depth information while our approach gives similar or better results than RGB based approaches.
The most important improvement of our approach is in terms of object tracking accuracy as we can see in table \ref{table:obj_pose}.
\begin{table*}
\caption{Camera pose estimation comparison on the Kitti tracking dataset.}
\vspace{-2mm}
\label{table:ego}
\centering
\resizebox{\textwidth}{!}{%
\begin{tabular}{@{}c|cc|cc|ll|cc|cc@{}}
\toprule
\multirow{2}{*}{seq} & \multicolumn{2}{c|}{ORB-SLAM2 \cite{mur2017orb}} & \multicolumn{2}{c|}{DynaSLAM \cite{bescos2018dynaslam}} & \multicolumn{2}{l|}{VDO-SLAM (RGB-D) \cite{zhang2020vdo}} & \multicolumn{2}{c|}{DynaSLAM2 \cite{bescos2021dynaslam}} & \multicolumn{2}{c}{Ours} \\ \cmidrule(l){2-11} 
 & RPE$_t$ (m/f) & RPE$_R$ (°/f) & RPE$_t$ (m/f) & RPE$_R$ (°/f) & RPE$_t$ (m/f) & RPE$_R$ (°/f) & RPE$_t$ (m/f) & RPE$_R$ (°/f) & RPE$_t$ (m/f) & RPE$_R$ (°/f) \\ \midrule
00 & \bf{0.04} & 0.06 & \bf{0.04} & 0.06 & 0.07 & 0.07 & \bf{0.04} & 0.06 & \bf{0.04} & \bf{0.05} \\
01 & 0.05 & 0.04 & 0.05 & 0.04 & \bf{0.04} & 0.12 & 0.05 & 0.04 & \bf{0.04} & \bf{0.03} \\
02 & 0.04 & 0.03 & 0.04 & 0.03 & \bf{0.02} & 0.04 & 0.04 & \bf{0.02} & 0.03 & 0.03 \\
03 & 0.07 & 0.04 & 0.07 & 0.04 & \bf{0.03} & 0.08 & 0.06 & 0.04 & 0.06 & \bf{0.02} \\
04 & 0.07 & 0.06 & 0.07 & 0.06 & \bf{0.05} & 0.11 & 0.07 & 0.06 & 0.06 & \bf{0.04} \\
05 & 0.06 & 0.03 & 0.06 & 0.03 & \bf{0.02} & 0.09 & 0.06 & 0.03 & 0.06 & \bf{0.02} \\
06 & \bf{0.02} & 0.04 & \bf{0.02} & 0.04 & 0.05 & 0.02 & \bf{0.02} & \bf{0.01} & \bf{0.02} & 0.04 \\
07 & 0.05 & 0.07 & 0.05 & 0.07 & - & - & 0.05 & 0.07 & \bf{0.04} & \bf{0.04} \\
08 & 0.08 & 0.04 & 0.08 & 0.04 & - & - & 0.10 & 0.04 & \bf{0.07} & \bf{0.03} \\
09 & 0.06 & 0.05 & 0.06 & 0.05 & - & - & 0.06 & 0.06 & \bf{0.05} & \bf{0.04} \\
10 & \bf{0.07} & 0.04 & \bf{0.07} & 0.04 & - & - & \bf{0.07} & \bf{0.03} & \bf{0.07} & \bf{0.03} \\
11 & 0.04 & 0.03 & 0.04 & 0.03 & - & - & 0.04 & 0.03 & \bf{0.03} & \bf{0.02} \\
13 & 0.04 & 0.05 & 0.04 & 0.05 & - & - & 0.04 & \bf{0.04} & \bf{0.03} & \bf{0.04} \\
14 & \bf{0.03} & 0.08 & \bf{0.03} & 0.08 & - & - & \bf{0.03} & 0.08 & \bf{0.03} & \bf{0.06} \\
18 & 0.05 & 0.03 & 0.05 & 0.03 & \bf{0.02} & 0.07 & 0.05 & \bf{0.02} & 0.04 & \bf{0.02} \\
19 & 0.05 & 0.03 & 0.05 & 0.03 & - & - & 0.05 & 0.02 & \bf{0.03} & \bf{0.03} \\
20 & 0.11 & 0.07 & 0.05 & 0.04 & \bf{0.03} & 0.17 & 0.07 & 0.04 & 0.04 & \bf{0.03} \\ \midrule
mean & 0.055 & 0.046 & 0.051 & 0.045 & - & - & 0.053 & 0.041 & \bf{0.044} & \bf{0.034} \\ \midrule
std & 0.020 & 0.016 & 0.016 & 0.015 & - & - & 0.019 & 0.020 & \bf{0.015} & \bf{0.011} \\ \bottomrule
\end{tabular}%
}
\end{table*}

\subsection{Object pose estimation.}
In this subsection we evaluate the accuracy of our object pose estimation. 
As we can see we improve object tracking accuracy, particularly for static objects such as the car 35 from sequence 11. The most important improvements usually come from the rotational part of the RPE, which is understandable as we only have 1 degree of freedom for the rotation of cars. 
We also observe that the most challenging cases happen when an object starts and stays far from the camera (e.g. seq. 05 and 10) because object tracking uses 3D points triangulated from stereo matches that are imprecise when points are far from the camera. We argue that the bruteforce keypoint matching of \cite{bescos2021dynaslam} help them when few frame to frame matches can be found, which can happen when the object is far from the camera.
Furthermore we have not implemented a way to relocalize an object that has been lost for multiple frames. Thus on some sequences (e.g. car 0 of seq. 11 and car 12 of seq. 20) in which the objects are alternatively far and close from the camera, we are only able to track them on a small portion of their trajectory.
However, we can see that we are generally able to accurately track objects for most of their trajectory. 
During our experiments we saw that we were able to track pedestrians, despite the fact that they are not rigid. We believe that our approach works because pedestrians undergo small deformations around arms and legs. 
\rev{As for camera pose estimation, we computed the standard deviation of object pose estimation for the sequence 3 and obtained values of $3.2\times 10^{-2}$, $2.8 \times 10^{-2}$ and $4.7\times 10^{-2}$ for the APE, the translational RPE and the rotational RPE. We also evaluated our results with an ANOVA which shows a significant difference (p-values $\leq$ 0.1) for most of our experiments.}
\begin{table*}
\caption{Object pose estimation comparison on the Kitti tracking dataset. ATE is in m, RPE$_t$ in m/m, RPE$_R$ in °/m, TP and MOTP in \%. }
\vspace{-2mm}
\label{table:obj_pose}
\resizebox{\textwidth}{!}{%
\begin{tabular}{@{}ccccccccccccccccccc@{}}
\toprule
 & \multicolumn{9}{c||}{DynaSLAM 2 \cite{bescos2021dynaslam}} & \multicolumn{9}{c}{TwistSLAM} \\ \midrule
\multicolumn{1}{c|}{seq / obj. id / class} & ATE & RPE$_t$ & \multicolumn{1}{c|}{RPE$_R$} & 2D TP & 2D MOTP & BV TP & BV MOTP & 3D TP & \multicolumn{1}{c||}{3D MOTP} & ATE & RPE$_t$ & \multicolumn{1}{c|}{RPE$_R$} & 2D TP & 2D MOTP & BV TP & BV MOTP & 3D TP & 3D MOTP \\ \midrule
\multicolumn{1}{c|}{03 / 1 / car}&0.69&0.34&\multicolumn{1}{c|}{1.84}&50.0&\textbf{71.79}&39.34&56.61&38.53&\multicolumn{1}{c||}{48.20}&\textbf{0.31}&\textbf{0.10}&\multicolumn{1}{c|}{\textbf{0.28}}& \textbf{58.02} & 60.00 & \textbf{58.02} & \textbf{60.00}  & \textbf{58.02} & \textbf{60.00}  \\
\multicolumn{1}{c|}{05 / 31 / car} & 0.51 & 0.26 & \multicolumn{1}{c|}{13.5} & 28.96 & \textbf{60.30} & 14.48 & \textbf{46.84} & 11.45 & \multicolumn{1}{c||}{34.20} & \textbf{0.35} & \textbf{0.19} & \multicolumn{1}{c|}{\textbf{0.58}} & \textbf{30.84} & 35.00 &\textbf{30.84} & 35.00 & \textbf{30.84} & \textbf{35.00} \\
\multicolumn{1}{c|}{10 / 0 / car} & 0.95 & 0.40 & \multicolumn{1}{c|}{2.84} & \bf{81.63} & \bf{73.51} & \bf{70.41} & \bf{47.60} & \bf{68.37} & \multicolumn{1}{c||}{\bf{40.28}} & \textbf{0.77} & \textbf{0.21} & \multicolumn{1}{c|}{\textbf{1.98}} & 7.20 & 3.70 & 6.10 & 3.10 & 5.80 & 2.80\\
\multicolumn{1}{c|}{11 / 0 / car} & 1.05 & 0.43 & \multicolumn{1}{c|}{12.51} & \bf{72.65} & \bf{74.78} & \bf{61.66} & \bf{50.74} & \bf{52.28} & \multicolumn{1}{c||}{\bf{47.35}} & \bf{0.17} & \bf{0.23} & \multicolumn{1}{c|}{\bf{0.23}} & 29.61 & 32.50 & 29.61 & 32.50 & 29.61 & 32.50 \\
\multicolumn{1}{c|}{11 / 35 car} & 1.25 & 0.89 & \multicolumn{1}{c|}{16.64} & 53.17 & 65.25 & 19.05 & 31.95 & 6.35 & \multicolumn{1}{c||}{26.02} & \textbf{0.10} & \textbf{0.03} & \multicolumn{1}{c|}{\textbf{0.11}} & \textbf{65.00} & \textbf{67.50} & \textbf{65.00} & \textbf{67.50} & \textbf{65.00} & \textbf{67.50} \\
\multicolumn{1}{c|}{18 / 2 / car} & 1.10 & 0.30 & \multicolumn{1}{c|}{9.27} & \textbf{86.36} & 74.81 & 67.05 & 45.47 & 62.12 & \multicolumn{1}{c||}{34.80} & \textbf{0.21} & \textbf{0.27} & \multicolumn{1}{c|}{\textbf{0.66}} & 84.67 & \textbf{87.50} & \textbf{84.67} & \textbf{87.50} & \textbf{84.67} & \textbf{87.50} \\
\multicolumn{1}{c|}{18 / 3 / car} & 1.13 & 0.55 & \multicolumn{1}{c|}{20.05} & \bf{53.33} & \bf{70.94} & 21.75 & \bf{41.45} & 16.84 & \multicolumn{1}{c||}{\bf{35.80}} & \bf{0.15} & \bf{0.21} & \multicolumn{1}{c|}{\bf{0.56}} & 28.19 & 30.00 & \bf{28.19} & 30.00 & \bf{28.19} & 30.00 \\
\multicolumn{1}{c|}{19 / 63 / car} & 0.86 & \bf{1.45} & \multicolumn{1}{c|}{48.80} & 35.26 & 63.50 & 29.48 & 45.69 & 26.48 & \multicolumn{1}{c||}{\bf{33.89}} & \bf{0.28} & 2.17 & \multicolumn{1}{c|}{\bf{1.08}} & 65.93 & 70.00 & 65.93 & 70.00 & 36.26 & 20.64 \\
\multicolumn{1}{c|}{19 / 72 / car} & 0.99 & 1.12 & \multicolumn{1}{c|}{3.36} & \bf{29.11} & \bf{62.59} & \bf{29.43} & \bf{55.48} & \bf{29.43} & \multicolumn{1}{c||}{39.81} & \bf{0.16} & \bf{0.05} & \multicolumn{1}{c|}{\bf{0.34}} & 16.92 & 20.00 & 16.92 & 20.00 & 16.92 & 20.00 \\
\multicolumn{1}{c|}{20 / 0 / car} & 0.56 & 0.45 & \multicolumn{1}{c|}{1.30} & 63.68 & 78.54 & 43.78 & 45.00 & 31.84 & \multicolumn{1}{c||}{46.15} & \bf{0.17} & \bf{0.20} & \multicolumn{1}{c|}{\bf{0.72}} & \bf{84.75} & \bf{87.50} & \bf{84.75} & \bf{87.50} & \bf{84.75} & \bf{87.50} \\
\multicolumn{1}{c|}{20 / 12 / car} & 1.18 & 0.40 & \multicolumn{1}{c|}{6.19} & \bf{42.77} & \bf{76.77} & \bf{37.64} & \bf{49.29} & \bf{36.23} & \multicolumn{1}{c||}{40.81} & \bf{0.24} & \bf{0.20} & \multicolumn{1}{c|}{\bf{1.54}} & 14.24 & 17.5 & 13.91 & 17.45 & 13.04 & 17.25 \\
\multicolumn{1}{c|}{20 / 122 / car} & 0.87 & 0.72 & \multicolumn{1}{c|}{5.75} & 34.90 & 78.76 & 34.51 & 48.05 & 29.02 & \multicolumn{1}{c||}{44.43} & \bf{0.17} & \bf{0.02} & \multicolumn{1}{c|}{\bf{0.07}} & \bf{84.94} & \bf{87.50} & \bf{84.94} & \bf{87.50} & \bf{84.94} & \bf{87.50} \\ \midrule
\multicolumn{1}{c|}{mean} & 0.93& 0.61 & \multicolumn{1}{c|}{11.83} & 55.15 & 70.96 & 39.05& 47.01  & 34.08 & \multicolumn{1}{c||}{39.31 } & \bf{0.26} & \bf{0.32} & \multicolumn{1}{c|}{\bf{0.68}} & \bf{45.53} & \bf{49.89} & \bf{47.41} & \bf{49.84} & \bf{44.84} & \bf{43.18} \\ \midrule
\multicolumn{1}{c|}{std} & 0.23 & 0.35 & \multicolumn{1}{c|}{12.58} & 17.65 & 6.19 & 17.82 &6.13 & 18.27 & \multicolumn{1}{c||}{6.36} & 0.18 & 0.58 & \multicolumn{1}{c|}{0.34} & 29.49 &30.24 & 29.66 & 30.32 & 29.33 & 30.02 \\ \bottomrule
\end{tabular}%
}
\end{table*}

\begin{figure*}
    \centering
    \includegraphics[width=0.98\textwidth]{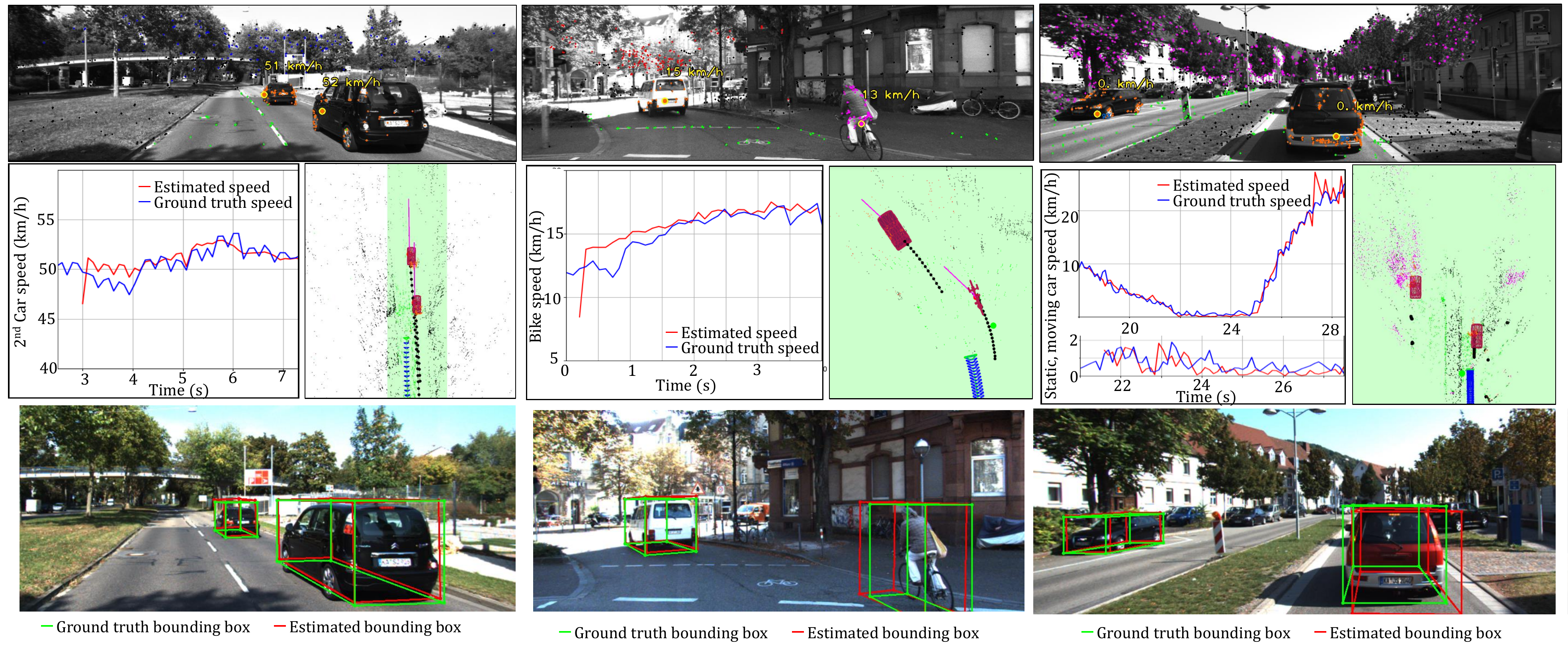}
    \caption{Qualitative examples of cluster tracking on sequences from the KITTI tracking dataset. (Top) Frame with tracked clusters and their speed. (Middle left) Comparison of estimated (red) and ground truth (blue) speed (in km/h). (Middle right) Map with tracked clusters and camera poses, seen from above. (Bottom) Visualization of estimated poses (red) and ground truth (green) represented by their 3D bounding boxes.}
    \label{fig:quali}
    \vspace{-5mm}
\end{figure*}

We also show some qualitative results for the mapping, the camera and object pose estimation. The results are visible in figure \ref{fig:quali}. We are able to track multiple objects on all sequences. The estimated speed is very close from the ground truth with a maximum difference of about 3 km/h which occurs when the object is far from the camera or when it is created. Looking at the bounding boxes we can see that they coincide and thus that the poses are well estimated for near and far objects.
As we can see in the middle figure we can accurately track non rigid objects such as the cyclist, as long as most of their surface is rigid. We also show on the right an example of both a tracked static car and a car that slows down and speeds up. As we can see the estimated speed is close to the 0 for the static car, making the dynamic keypoints act like static ones. 
\section{Conclusion}
In this paper we proposed a new stereo semantic dynamic SLAM system called TwistSLAM, able to estimate both the pose of the camera as well as to track all dynamic objects in the scene. Using mechanical joints between clusters we can constrain objects movements to physically possible movements, which allows us to improve both camera and objects pose estimation compared to the state of the art.




\bibliographystyle{IEEEtran} 
\bibliography{IEEEabrv,IEEEexample}
\end{document}